\documentclass[sigconf, screen = False]{acmart}
\usepackage{graphicx}
\usepackage{subcaption}
\usepackage{tabularx}
\usepackage{adjustbox}
\newcolumntype{"}{@{\hskip\tabcolsep\vrule width 1pt\hskip\tabcolsep}}
\makeatother
\AtBeginDocument{%
  \providecommand\BibTeX{{%
    \normalfont B\kern-0.5em{\scshape i\kern-0.25em b}\kern-0.8em\TeX}}}

\setcopyright{acmcopyright}
\copyrightyear{2023}
\acmYear{2023}
\acmDOI{XXXXXXX.XXXXXXX}

\acmConference[CIKM '23]{The 32nd ACM International Conference on Information and Knowledge Management}{October 21-25, 2023}{Birmingham, UK}
\acmPrice{\textcolor{red}{15.00}}
\acmISBN{\textcolor{red}{978-1-4503-XXXX-X/18/06}}

\begin{document}

\title{MC-DRE: Multi-Aspect Cross Integration\\ for Drug Event/Entity Extraction}

\author{Jie Yang}
\email{jyan4704@uni.sydney.edu.au}
\affiliation{%
\institution{The University of Sydney}
\country{Australia}
}

\author{Soyeon Caren Han}
\authornote{Corresponding author}
\email{caren.han@uwa.edu.au}
\affiliation{%
\institution{The University of Western Australia, The University of Sydney} 
\country{Australia}
}

\author{Siqu Long}
\email{slon6753@uni.sydney.edu.au}
\affiliation{%
\institution{The University of Sydney} 
\country{Australia}
}

\author{Josiah Poon}
\email{josiah.poon@sydney.edu.au}
\affiliation{%
\institution{The University of Sydney} 
\country{Australia}
}

\author{Goran Nenadic}
\email{gnenadic@manchester.ac.uk}
\affiliation{%
\institution{The University of Manchester} 
\country{United Kingdom}
}


\begin{abstract}
Extracting meaningful drug-related information chunks, such as adverse drug events (ADE), is crucial for preventing morbidity and saving many lives. Most ADEs are reported via an unstructured conversation with the medical context, so applying a general entity recognition approach is not sufficient enough. In this paper, we propose a new multi-aspect cross-integration framework for drug entity/event detection by capturing and aligning different context/language/knowledge properties from drug-related documents. We first construct multi-aspect encoders to describe semantic, syntactic, and medical document contextual information by conducting those slot tagging tasks, main drug entity/event detection, part-of-speech tagging, and general medical named entity recognition. Then, each encoder conducts cross-integration with other contextual information in three ways: the key-value cross, attention cross, and feedforward cross, so the multi-encoders are integrated in depth. Our model outperforms all SOTA on two widely used tasks, flat entity detection and discontinuous event extraction.\footnote{The code can be found at~\url{https://github.com/adlnlp/mc-dre}.}
\end{abstract}






\maketitle

\section{Introduction}
The World Health Organization\footnote{\url{https://www.who.int/docs/default-source/medicines/safety-of-medicines--adverse-drug-reactions-jun18.pdf?sfvrsn=4fcaf40_2}} defines Adverse Drug Reactions (ADR) as `Harmful, unintended reactions to medicines that occur at doses normally used for treatment'. According to the WHO, most ADE-related information and related medications are reported via an unstructured conversation, such as electronic health records or social media, with a medical domain context. Hence, a full understanding of the unstructured sequence of slot values with in-depth medical domain expertise is crucial. Due to this nature, general entity/token tagging frameworks are insufficient for drug-related entity(DE) extraction. Existing DE extraction research studies have adopted relevant multiple aspects, such as pre-trained health informatics contextual embedding and external medical knowledge bases. To handle this multi-aspect data and combine its information, DE extraction researchers have applied mainly two types of multi-aspect fusion techniques\cite{weld2021conda,wang2020detect,han21f_interspeech}, including early (data-level) fusion \cite{chen2020extracting,ju2020ensemble,dandala2020extraction,wei2020study} and late (decision-level) fusion \cite{dai2020adverse,kim2020ensemble,yang2023ddi}. Early fusion techniques focus on combining all individual medical input aspects into a unified representation before proceeding through the learning. A clear common representation of all multi-aspects is crucial so the specific and unique properties of each aspect may be lost by an early fusion. No doubt that such information leakage would affect the proper medical/drug entity extraction performance. On the other hand, late fusion aims to compute separately for each feature and concatenated thereafter. During late fusion, direct interaction effects between multiple medical aspects are lacking and tend to ignore some low-level interaction since it is not possible to update the cost function of the multiple aspect-based models. An ideal fusion for successful drug-related entity extraction would synergistically combine multiple aspects and ensure the resultant product reflects the salient features of different drug-related aspects.

In this paper, we propose a novel multi-aspect cross integration framework for drug entity detection by capturing and aligning different context/language/knowledge properties from drug-related documents. To achieve this aim, we first select three aspects, semantic information of drug-related slot tokens, syntactic structure information of a medical text, and medical domain expertise. Based on such aspects, we construct multi-aspect encoders to describe semantic, syntactic, and general medical contextual information by conducting three different tasks, including main drug entity detection, part-of-speech tagging, and general medical named entity recognition. Each encoder conducts cross-fusion techniques to jointly integrate and align with other contextual information. We apply and validate three different cross-fusion techniques, 1) key-value cross, 2) attention cross, and 3) feedforward cross, so the multi-encoders are integrated into depth. 

Our Contributions are as follows: 1) We propose a new multi-aspect cross-integration framework for drug-related entity extraction that enables synergistic integration and alignment of different context/language/knowledge properties. 2) Our model (MC-DRE) outperforms all twelve state-of-the-art models from both Flat and Discontinuous Drug-related Entity Extraction tasks.

\begin{figure}[t]
    \centering
    \includegraphics[width = \columnwidth]{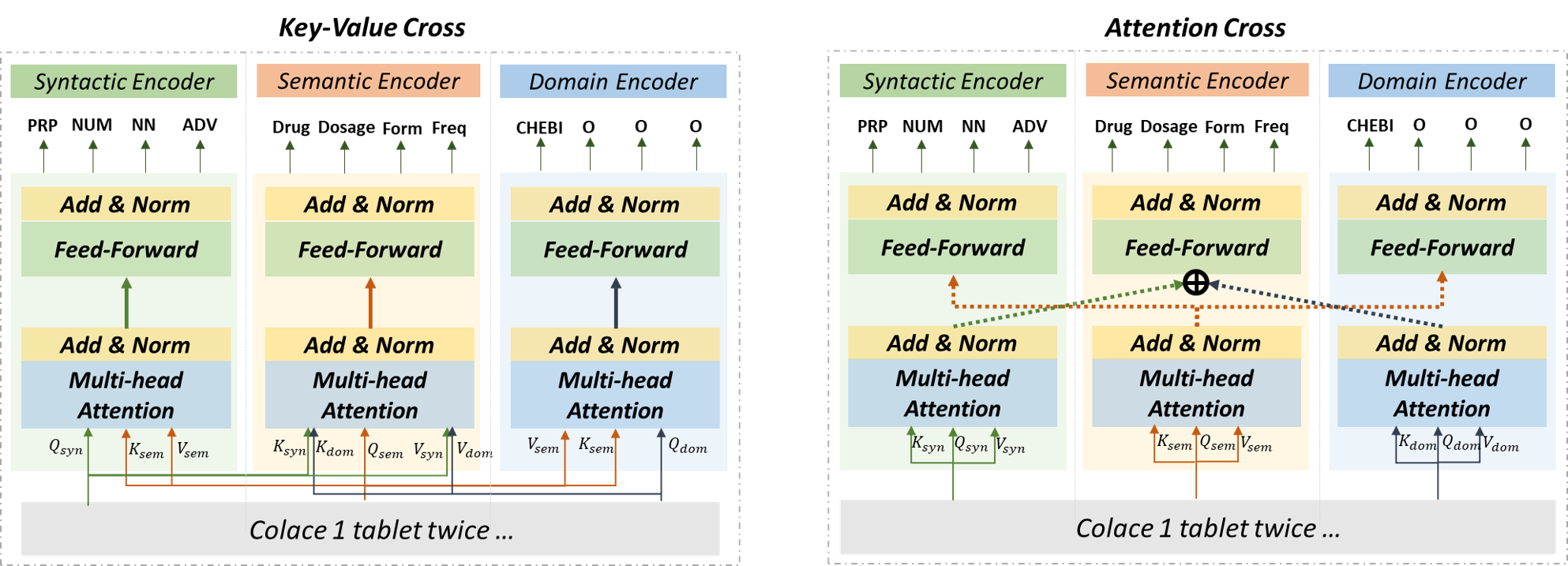}
    \caption{Overview of Multi-aspect Cross Integration Framework. The key-value cross (left) and attention cross (right) integration mechanisms are illustrated considering their outstanding performance (See Section 4.2).}    
    \label{fig:ArchiOverall}
    \Description{TODO}
\end{figure}

\section{MC-DRE}
In this study, we utilize the transformer encoder as the fundamental component to gather multiple aspects of information. We present details of the input embedding, how to construct the multi-aspect encoders, and the cross-integration learning and the prediction. 

\textbf{Contextual Input Representation}
Recently, deep learning architectures with pre-trained language models (PLMs) could achieve state-of-the-art (SOTA) performance on various general domain NLP tasks, because contextualized embeddings from PLMs represent different meanings based on the context (e.g, \textit{date} can mean \textit{time}, \textit{a kind of fruit}, and \textit{appointment} in disparate texts). More specifically, some words may have opposite meanings in the medical domain (e.g, \textit{positive} usually means something good, but often refers to the presence of a specific condition, which is typically not a desirable outcome). Inspired by this, we explored several medical-specific PLMs, i.e., BioBERT \cite{lee2020biobert}, ClinicalBERT \cite{huang2019clinicalbert}, and PubMedBERT \cite{GuPubMedBERT}, and found that PubMedBERT would be the best performed \textit{`medical'} contextual embeddings\footnote{The input embedding ablation studies are already done and will be included in the Appendix after the acceptance.}.

\begin{table*}[t]

\caption{Overall Performance on 2018 n2c2 with Breakdown F-scores for each entity category. All baselines applied the same setup as ours, in terms of dataset split and evaluation metrics.}
\small
\begin{adjustbox}{width = \textwidth}
% \resizebox{\columnwidth}{!}{
\addtolength\tabcolsep{1pt}
\begin{tabular}{lcccccccccc} 
\specialrule{1pt}{0pt}{0pt}
\textbf{Models}&
  \multicolumn{1}{l}{\textbf{ADE}} &
  \multicolumn{1}{l}{\textbf{Dosage}} &
  \multicolumn{1}{l}{\textbf{Drug}} &
  \multicolumn{1}{l}{\textbf{Duration}} &
  \multicolumn{1}{l}{\textbf{Form}} &
  \multicolumn{1}{l}{\textbf{Frequency}} &
  \multicolumn{1}{l}{\textbf{Reason}} &
  \multicolumn{1}{l}{\textbf{Route}} &
  \multicolumn{1}{l}{\textbf{Strength}} &
  \multicolumn{1}{l}{\textbf{Overall}}\\ 
  
 &
  \multicolumn{1}{l}{\textbf{2\%}} &
  \multicolumn{1}{l}{\textbf{8\%}} &
  \multicolumn{1}{l}{\textbf{32\%}} &
  \multicolumn{1}{l}{\textbf{1\%}} &
  \multicolumn{1}{l}{\textbf{13\%}} &
  \multicolumn{1}{l}{\textbf{12\%}} &
  \multicolumn{1}{l}{\textbf{8\%}} &
  \multicolumn{1}{l}{\textbf{11\%}} &
  \multicolumn{1}{l}{\textbf{13\%}} &
  \multicolumn{1}{l}{\textbf{100\%}} 
\\

  \hline
Chen et al. \cite{chen2020extracting} &
  42.08\% &
  88.37\% &
  86.92\% &
  70.11\% &
  88.99\% &
  90.11\% &
  57.84\% &
  86.73\% &
  92.62\% &
  84.97\% \\
Dai et al. \cite{dai2020adverse}&
  38.75\% &
  92.67\% &
  93.10\% &
  81.61\% &
  94.94\% &
  96.95\% &
  62.67\% &
  94.70\% &
  97.41\% &
  91.90\% \\
Kim et al. \cite{kim2020ensemble}&
  27.11\% &
  93.93\% &
  95.55\% &
  81.78\% &
  95.46\% &
  97.06\% &
  57.57\% &
  95.40\% &
  97.88\% &
  92.66\% \\
Ju et al. \cite{ju2020ensemble} &
  27.90\% &
  93.95\% &
  95.50\% &
  81.38\% &
  95.43\% &
  97.27\% &
  62.37\% &
  95.52\% &
  98.10\% &
  92.78\% \\
Dandala et al. \cite{dandala2020extraction} &
  46.20\% &
  94.10\% &
  95.40\% &
  83.50\% &
  \underline{95.80\%} &
  97.00\% &
  67.60\% &
  95.30\% &
  97.40\% &
  92.90\% \\
Narananan et al. \cite{narayanan2020evaluation}&
  53.08\% &
  93.22\% &
  94.75\% &
  85.65\% &
  95.57\% &
  97.10\% &
  \underline{68.57\%} &
  95.38\% &
  97.97\% &
  92.91\% \\
Wei at al. \cite{wei2020study} &
  52.95\% &
  \underline{94.82\%} &
  \underline{95.56\%} &
  \underline{86.24\%} &
  95.75\% &
  \underline{97.48\%} &
  67.49\% &
  \underline{95.62\%} &
  \underline{98.32\%} &
  93.45\% \\
% \cite{el2021mttlade}(2021) &

%   \textbf{69.39\%} &
%   \textbf{94.86\%} &
%   95.03\% &
%   86.13\% &
%   95.62\% &
%   95.58\% &
%   \underline{81.93\%} &
%   94.65\% &
%   96.32\% &
%   93.65\% \\
Narayanan et al.\cite{NARAYANAN2022103960}&

  \underline{56.04\%} &
  91.43\% &
  93.39\% &
  77.66\% &
  93.11\% &
  86.11\% &
  65.41\% &
  94.13\% &
  95.74\% &
  \underline{94.00\%} \\
  \textbf{Ours} &

  \textbf{61.59\%} &
  \textbf{98.11\%} &
  \textbf{98.86\%} &
  \textbf{93.76\%} &
  \textbf{98.80\%} &
  \textbf{98.84\%} &
  \textbf{87.13\%} &
  \textbf{98.41\%} &
  \textbf{99.43\%} &
  \textbf{98.38\%} \\ 
\specialrule{1pt}{0pt}{0pt}

\end{tabular}
\end{adjustbox}
\label{tab:breakdown_n2c2}
\end{table*}
\subsection{Multi-aspect Encoders}
Both flat drug entity and discontinuous adverse drug event extractions are token-level tasks, which give a possible label for each token in a sequence. Due to the complexity of medical texts, it is necessary to understand the meaning of each token and the grammatical and syntactic structure as tokens have different aspects. Also, overall medical documents would include lots of special terminologies and abbreviations so understanding those terms may benefit the final prediction. Thus, we use multi-aspect encoders to obtain drug event semantic, syntactic, and medical document contextual information by leading different medical slot tagging tasks. Details of the multi-encoders are presented as follows:

\textbf{Semantic Encoder}
Drug entity detection is one of the sub-tasks of medical spoken language understanding (MSLU), which seeks to extract semantic elements from input sentences. Inspired by previous studies \cite{narayanan2020evaluation,el2021mttlade}, we feed the output of the input embedding module into an encoder to acquire semantic contextual knowledge.

\textbf{Syntactic Encoder}
Part-of-speech (POS) tagging is the process of identifying and labelling the parts of speech in a sentence. This process aids the model in understanding the syntactic structure. In this paper, we extract POS tags from two libraries, i.e., spaCy\footnote{https://spacy.io/} and NLTK\footnote{https://www.nltk.org/}, and feed them into an encoder to explore the syntactic information. Similar to the strategies in the semantic encoder, we utilise the word representation from the input embedding module and obtain syntactic-based contextual information as the output.

\textbf{Domain-Specific Encoder}
General medical named entities include such as diseases, treatments, DNA, RNA, protein, and other medical concepts. The output can support other medical applications, like clinical information retrieval, and disease surveillance. Here, based on the contextual embedding from input layer, we explore four kinds of models in scispaCy \cite{neumann-etal-2019-scispacy}, and feed them into an encoder to collect the medical entity-based contextual knowledge.

After obtaining multiple crucial aspects of information by the above encoders, we focus on exploiting joint learning approaches to integrate and align information among encoders.

\subsection{Cross Learning and Integration}
We propose three cross-learning methods to jointly integrate and align the above multi-aspect encoders. Despite the identical input in the encoders, the contextual knowledge can be updated via the backpropagation algorithm within the neural network architecture. 

\textbf{1) key-value input cross} The concept of key-value input cross involves sharing the K and V matrices of the multi-head attention in any two encoders with the third ones. Take semantic (se.) encoder as the example, the original K and V, i.e., $K_{se.}$ and $V_{se.}$, will be updated by the concatenation of K and V in the syntactic (sy.) and domain-specific (do.) encoders. Then we feed the attention output into a dense layer to adjust the tensor dimensions.
\begin{equation}
K_{se.new} = concat(K_{sy.}, K_{do.}), 
V_{se.new} = concat(V_{sy.}, V_{do.})
\end{equation}
\begin{equation}
c_i = attention(Q_{se.}W_i^{Q_{se.}}, K_{se.new}W_i^{K_{se.new}}, V_{se.new}W_i^{V_{se.new}})
\end{equation}
 
\textbf{2) attention cross} The knowledge after the multi-head attention in one encoder will be updated based on the attention score in the other two. By concatenating the attention output from the sy. and do. encoders, we will update the attention score in se.
% \begin{equation}
% MultiHead_{se.new} = concat(MultiHead_{sy.}, MultiHead_{do.})
% \end{equation}

\textbf{3) feedforward cross}
The formula for the layer normalization after the FFN layer in the transformer encoder is written as: 
\begin{equation}
LayerNorm(x + FFN(x))
\end{equation}
where \textit{x} is the input to the FFN layer as introduced in formula 3
For this integration method, information after the FFN sub-layer, i.e., \textit{FFN(x)}, will be shared among the three encoders, so that we calculate the layer normalization in se. encoder as:
\begin{equation}
LayerNorm(x_{se.} + concat(FFN(x_{sy.}, FFN(x_{do.}))))
\end{equation}
where $FFN(x_{sy.})$ and $FFN(x_{do.})$ are the output of FFN sub-layer in sy. and do. encoders, respectively. 
Like semantic encoders, syntactic and domain-specific encoders also fused information from the other two. Let $J_{se.}$, $J_{sy.}$, and $J_{do.}$ represent the output from joint integration learning for the three encoders, respectively. 

\textbf{Drug Entity Extraction}
Finally, we perform softmax to give probability distributions over the drug entity, POS, and general medical NER labels on each token in the three encoders, separately:  
\begin{equation}
P_{se.} = softmax(J_{se.} * W^{se.} + b^{se.})
\end{equation}
\begin{equation}
P_{sy.} = softmax(J_{sy.} * W^{sy.} + b^{sy.})
\end{equation}
\begin{equation}
P_{do.} = softmax(J_{do.} * W^{do.} + b^{do.})
\end{equation}
The multi-aspect cross integration model is trained on the sum of the cross entropy losses for the three tasks.

\section{Evaluation Setup}
Two popular public benchmark datasets 
%for flat drug entity extraction and discontinuous ADE extraction 
were applied. \textbf{1) Flat Drug Entity: } The second track in the 2018 National NLP Clinical Challenge shared task (2018 n2c2)\footnote{https://portal.dbmi.hms.harvard.edu/projects/n2c2-nlp/} \cite{Henry20202018NS} focused on the extraction of flat drug-related entities, including \textit{drugs}, their attributes (\textit{strength, form, frequency, route, dosage, reason, ADE} and \textit{duration}). Baselines are \cite{chen2020extracting,dai2020adverse,kim2020ensemble,ju2020ensemble,dandala2020extraction,narayanan2020evaluation,wei2020study,NARAYANAN2022103960}.
\textbf{2) Discontinuous Drug Entity: }
The CSIRO Adverse Drug Event Corpus(CADEC)\footnote{https://data.gov.au/dataset/ds-dap-csiro\%3A10948/details?q=} \cite{karimi2015cadec} is source from AskaPatient\footnote{https://www.askapatient.com/}. The entity types in CADEC contain \textit{drug}, \textit{disease, symptom}, and \textit{ADE} from 1,250 posts for 12 kinds of drugs {\textit{Voltaren, Catafam, Voltaren-XR, Arthrotec, Pennsaid, Solaraze, Flector, Cambia, Zipsor, Diclofenac Sodium, Diclofenac Potassium}, and \textit{Lipitor}}. To compare our performances directly against previously discontinuous ER models, like \cite{dai2020-effective,yan-etal-2021-unified-generative}, only the 1,000 annotated ADE are used in this paper. We follow \cite{TangBuzhou2018RCaD} to extend the standard NER schema `BIO' (B-beginning of an entity, I-inside of an entity, O-outside of an entity) into `BIOHD'. `H' means the token is shared by multiple mentions, and `D' means a discontinuous entity but not shared by other mentions \cite{TangBuzhou2018RCaD}. Baselines are \cite{dai2020-effective,yan-etal-2021-unified-generative,zhang2022bias,li2022unified}.

\textbf{Evaluation Metrics} \textit{F-score} is the major evaluation metric in drug entity recognition. For the fair comparison with baselines, we adopt the \textit{lenient} and \textit{strict} micro-average F-score to evaluate the performance on the 2018 n2c2 and CADEC datasets, respectively. The lenient evaluation mode is applied to the 2018 n2c2 dataset. It allows an overlapped boundary between the gold annotation and prediction. The strict mode is applied to the CADEC dataset, followed by the study of \cite{TangBuzhou2018RCaD,dai2020-effective,yan-etal-2021-unified-generative}, which needs the prediction boundary with the exact match of the gold annotation. 

\textbf{Implementation settings}\footnote{We followed the data split and environment as the same as the original data papers \cite{Henry20202018NS,karimi2015cadec}  published.}
Regarding the hardware setting of the experiments, we use Google Colaboratory (Colab) as the development platform. The utilized deep learning framework was Keras \cite{ketkar2017introduction}, which was built on top of Tensorflow \cite{45166}. In detail, we set the learning rate to 4e-4, batch size 32, dropout rate 0.5, optimizer as \textit{Adam}, and loss function as \textit{Sparse categorical crossentropy}.

\section{Evaluation Results}

% \footnote{ `-' indicates the absence of reported results from the paper.}

\begin{table}[t]
\caption{Overall Performance on CADEC (F-score). Following \cite{dai2020-effective}, we add the breakdown F-score for both sentences containing at least one Discontinuous Mention(2nd column) and Discontinuous Mentions only(3rd column).}
\centering
\begin{adjustbox}{width = 0.9\columnwidth}
% \resizebox{\columnwidth}{!}{
\addtolength\tabcolsep{1pt}
\begin{tabular}{p{0.32\columnwidth}ccc}
\specialrule{1pt}{0pt}{0pt}
% \multicolumn{3}{|l|}{\multirow{3}{*}{char literals}}
\multicolumn{1}{l}{\textbf{Models}} & \multicolumn{1}{l}{\textbf{Sent. with DM}} & \multicolumn{1}{l}{\textbf{DM only}} 
& \multicolumn{1}{l}{\textbf{Overall}}
\\
\hline

Dai et al. \cite{dai2020-effective}& 65.40\% & 37.90\%& 69.00\% \\ 
Yan et al. \cite{yan-etal-2021-unified-generative}&- &-   & 70.64\% \\ 
Zhang et al. \cite{zhang2022bias}&-&-&71.60\%\\ 
Li et al. \cite{li2022unified} &-&-&\underline{73.21\%} \\ 
\textbf{Ours} & \textbf{68.86\%} & \textbf{40.00\%} & \textbf{76.50\%} \\
\specialrule{1pt}{0pt}{0pt}
\end{tabular}
\end{adjustbox}
\label{tab:CADEC_overall}
\end{table}

\subsection{Overall Performance}
We compared the performance of our multi-aspect cross-integration framework with baselines. In Table~\ref{tab:breakdown_n2c2}, compared to those baselines on single-aspect or multi-aspect fusion, our proposed multi-aspect cross-integration mechanism achieved the best performance on all the Entity tags and produced the highest overall F-score of 98.38\% on 2018 n2c2, surpassing the second best by a large margin of 4.38\%. More specifically, improvements compared to the best baseline on each Entity type ranging from 1.11\% (Strength) up to 18.56\% (Reason) are observed, with a notable increase of 7.52\% on Duration (in spite of its 1\% supporting training samples) and 18.56\% on Reason respectively. Similarly, our multi-aspect cross-integration framework also illustrates the significant improvement of the overall F-score on CADEC, outperforming the best baseline by around 3.29\%. Following \cite{dai2020-effective}, we also evaluated for sentences with at least one discontinuous mentions and those containing discontinuous mentions only on CADEC. The results show an increase of 3.46\% and 2.1\% respectively, demonstrating the effectiveness of our framework in recognizing discontinuous mentions.

\subsection{Ablation Studies}
\textbf{Effect of Multi-aspect Encoders}
To evaluate the benefits of our proposed multi-aspect encoder, we conducted ablation studies on the two auxiliary aspects (i.e., Syntactic and Domain-based) for cross-learning. In Table~\ref{tab:ab_aspects}, the model with three-aspect encoders achieves the best performance on both datasets, 98.38\% on 2018 n2c2 and 76.50\% on CADEC. Besides, removing any of the two auxiliary aspects leads to a performance drop: 1) Without using the syntactic or domain aspect alone, around 2.53\% or 0.76\% decrease on 2018 n2c2 and around 3.56\% or 2.82\% drop on CADEC were found. 2) ablating both syntactic and domain aspects (i.e., single transformer modelling of one aspect without cross-learning of multi-aspect integration) leads to a more severe performance downgrade on both datasets. In summary, both datasets demonstrated more reliance on the syntactic aspect compared to the domain aspect whereas the two auxiliary aspects together boost the most with our multi-aspect cross-integration framework.

\noindent\textbf{Effect of Cross-integration Mechanism}
We ablated the cross-integration component by exploring three exchange mechanisms as it plays a key role in our research. Table~\ref{tab:ab_exchange} shows that simply modelling the three aspects via multi-task joint loss with \textit{no exchange} of aspect information produces inferior overall F-score on both datasets, i.e., 97.51\% for 2018 n2c2 and 75.05\% for CADEC. 
Different locations of exchanges enforce different degrees of aspect integration, resulting in performance gain variances on different datasets. For instance, the \textit{key-value cross} fuses the aspects at the earliest stage from the dense word-word relation modelling via attention components, while the \textit{attention cross} exchanges the aspects after the attention module and exerts the aspect fusion through the Feedforward layer and onwards. Comparatively, the \textit{feedforward cross} conducts the latest fusion only at the last Add\&Norm layer. As in Table~\ref{tab:ab_exchange}, each dataset demonstrates different preferences of exchanges that 2018 n2c2 performs the best with \textit{attention cross} (98.38\%) while CADEC prefers the comparative earlier cross via \textit{key-value cross} (76.50\%). We assume that the task of CADEC is more complex than 2018 n2c2 in terms of the number of Entity tags and the nature of discontinuousness, which benefits from more dense modelling of the multiple aspects at the early stage. 

\begin{table}[t]
\caption{Effect of Cross-integration Aspect (F-score). All variants applied the same aspect settings as the full multi-aspect cross-integration counterpart regarding the Semantic, Syntactic and Domain sources: PubMedBERT + NLTK + craft with attention cross (2018 n2c2) and PubMedBERT + spaCy + craft  with key-value cross (CADEC).}
\centering
\begin{adjustbox}{width = 0.9\columnwidth}
% \resizebox{\columnwidth}{!}{
\addtolength\tabcolsep{1pt}
\begin{tabular}{p{0.13\columnwidth}p{0.13\columnwidth}p{0.13\columnwidth}cc} 
\specialrule{1pt}{0pt}{0pt}
%\begin{tabular}{lllrr}
\textbf{Semantic} &
  \textbf{Syntactic} &
  \textbf{Domain} &
  \textbf{2018 n2c2} &
  \textbf{CADEC} \\\hline
o & x & x & 95.07\% & 71.34\% \\
o & o & x & \underline{97.62\%} & \underline{73.68\%} \\
o & x & o & 95.85\% & 72.94\% \\
o & o & o & \textbf{98.38\%} & \textbf{76.50\%} \\ 
\specialrule{1pt}{0pt}{0pt}

\end{tabular}
\label{tab:ab_aspects}
\end{adjustbox}
\end{table}

\begin{table}[t]
\caption{Effect of Cross-Integration Mechanism (F-score). }
\centering
\begin{adjustbox}{width = 0.9\columnwidth}
% \resizebox{\columnwidth}{!}{
\addtolength\tabcolsep{1pt}
\begin{tabular}{p{0.5\columnwidth}cc}
\specialrule{1pt}{0pt}{0pt}
  {\textbf{Cross learning approaches}} & {\textbf{2018 n2c2}} &{\textbf{CADEC}} \\ \hline
no exchange& 97.51\%  & 75.05\% \\
key-value input cross& 98.04\% & \textbf{76.50\%} \\
attention cross & \textbf{98.38\%} & \underline{75.82\%} \\
feedforward cross& \underline{98.26\%} & 74.83\% \\

\specialrule{1pt}{0pt}{0pt}  

\end{tabular}
\end{adjustbox}
\label{tab:ab_exchange}
\end{table}

\subsection{Qualitative Analysis: Case Study}
\begin{figure}[h]
    \centering
    \includegraphics[width = 0.7\columnwidth]{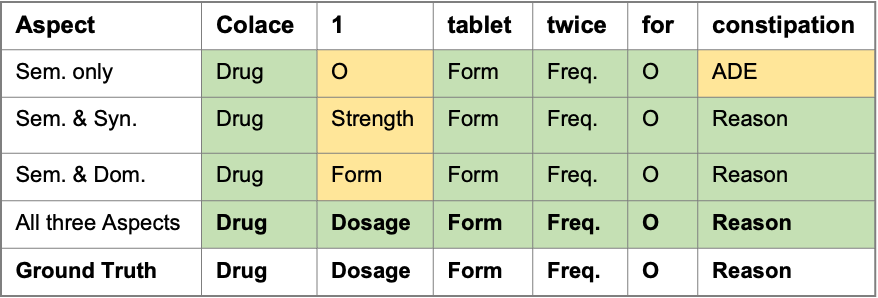}
    \caption{Prediction example with our model on 2018 n2c2.}    
    \label{fig:casestudy}
\end{figure}

We further evaluate MC-DRE with a qualitative assessment of the entity detection on 2018 n2c2. In Figure~\ref{fig:casestudy}, assume we have a sentence \textit{`Colace 1 tablet twice for constipation'} and test the drug entity detection by using our models with different aspect cross integration shown in Table~\ref{tab:ab_aspects}. 
First, we found that most variants misclassified the word \textit{`1'}, except the one with all three aspects combination. This trend presents the effectiveness of integrating all three aspects in order to detect the meaning of ambiguous terms. Secondly, the model with a sem. only predicted \textit{`constipation'} as an ADE due to the missed grammatical connection or general medical context, especially with the word \textit{`for'}. Unsurprisingly, the model with three encoders correctly predicted all labels for this case.

\section{Conclusion}
Drug event/entity extraction is an essential but challenging task in the medical field due to its unstructured nature and in-depth medical expertise.
We propose MC-DRE, which enables the synergistic integration and alignment of different context, language, and knowledge properties for drug-related entity extraction. The results indicate that our proposed framework, MC-DRE, is effective in integrating and aligning crucial aspects of drug event information. It is hoped that our model, MC-DRE, provides great insight into the effectiveness of multi-aspect cross-integration and representation in the future direction of such drug-related entity extraction tasks.

% \section*{Ethical Consideration}
% This study was reviewed and approved by the ethics review committee of the authors' institution and conducted in accordance with the principles of the Declaration.

% \begin{acks}
% \end{acks}

\bibliographystyle{ACM-Reference-Format}
\bibliography{mcdre}

% \appendix

\end{document}